\pdfoutput=1
\ifdefined\pdfminorversion\pdfminorversion=7\fi
\ifdefined\pdfsuppresswarningpagegroup\pdfsuppresswarningpagegroup=1\fi
\PassOptionsToPackage{table,dvipsnames}{xcolor}
\documentclass[11pt]{article}

\usepackage{acl}

 \usepackage{microtype}

\usepackage[english]{babel} % English as the main language.
\usepackage{tgtermes}

\usepackage{rotating}
\usepackage{graphicx}
\usepackage{amsmath}
\usepackage{amsfonts}
\usepackage{booktabs}
\usepackage{multirow}
\usepackage{subcaption}
\usepackage{import}
\usepackage{lipsum}
\usepackage[linesnumbered, boxed]{algorithm2e}
\usepackage{makecell}
\usepackage{amssymb}
\usepackage{fontawesome}
\usepackage{xcolor}

\usepackage{blindtext}
\usepackage{url}
\usepackage[T1]{fontenc}
\usepackage{pgfplots}
\pgfplotsset{compat=1.18}
\usepgfplotslibrary{groupplots}

\definecolor{acceptedblue}{HTML}{1f77b4}
\definecolor{rejectedred}{HTML}{d62728}
\definecolor{sciDarkBlue}{HTML}{0072B2}  % acl-guidelines-junior
\definecolor{sciSkyBlue}{HTML}{56B4E9}   % acl-guidelines-senior
\definecolor{sciGreen}{HTML}{009E73}     % ai-generated
\definecolor{sciVermilion}{HTML}{D55E00} % default
\definecolor{sciPurple}{HTML}{CC79A7}    % simple
\definecolor{rejectedorange}{HTML}{f58518}

\usepackage{tcolorbox}
\usepackage{spverbatim}
\usepackage{xcolor}
\usepackage{fancyvrb}
\usepackage{fvextra}
\usepackage{array}
\usepackage{todonotes}
\usepackage{booktabs}

\usepackage{enumitem}
\usepackage{tabularx}
\title{
\raisebox{-0.2\height}{\includegraphics[width=0.6cm]{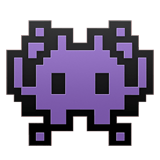}} 
Review Arcade:\\ On the Human Alignment and Gameability of LLM Reviews}
\author{
 \textbf{Hans Ole Hatzel\textsuperscript{1*}},
 \textbf{Sebastian Steindl\textsuperscript{3*}},
 \textbf{Jan Strich\textsuperscript{1,2*}}
 \\
 \textsuperscript{1}Language Technology Group, University of Hamburg, Germany \\
 \textsuperscript{2}Hub of Computing and Data Science (HCDS), University of Hamburg, Germany \\
 \textsuperscript{3}OTH Amberg-Weiden, Germany 
 \\
 \small{\textsuperscript{*}Equal contributions, order decided by coin toss.} \\   
 \small{\textbf{Correspondence:}} \\
 \small{
   \texttt{\{first\_name\}.\{last\_name\}@uni-hamburg.de},    \texttt{s.steindl@oth-aw.de}}
}

\begin{document}
\maketitle

\begin{abstract}
LLM-generated reviews for scientific papers are gaining considerable traction and are even being officially piloted by major conferences.
We have to assume that not only reviewers are using LLM-assistance, but also that authors use LLMs to revise their papers before submitting.
In this work, we perform empirical experiments on accepted and rejected submissions to the 2024--2025 cycles of the ACL Rolling Review (ARR) to evaluate LLM reviews from both the author and the reviewer perspective.
First, we identify a limited alignment of LLM reviews with human ones.
In the best-case scenario, the alignment is reasonable.
However, we also find that LLM-human alignment varies substantially across prompts and models.
Finally, we investigate the scenario in which the author uses an iterative draft-revise workflow to improve the submission according to the LLM review.
We find that this ``gaming'' of LLM reviews can be effective in specific scenarios, increasing overall scores for 36\% of the 984 papers.\par\smallskip
{\centering\mbox{%
\faGithub~\href{https://github.com/uhh-hcds/reviewarcade}{Code}\hspace{1.5em}%
\faDatabase~\href{https://huggingface.co/datasets/G4KMU/review_arcade}{Dataset}}\par}
\end{abstract}

\section{Introduction}\label{sec:intro}

\begin{figure}[!t]
    \centering
    \includegraphics[width=\linewidth]{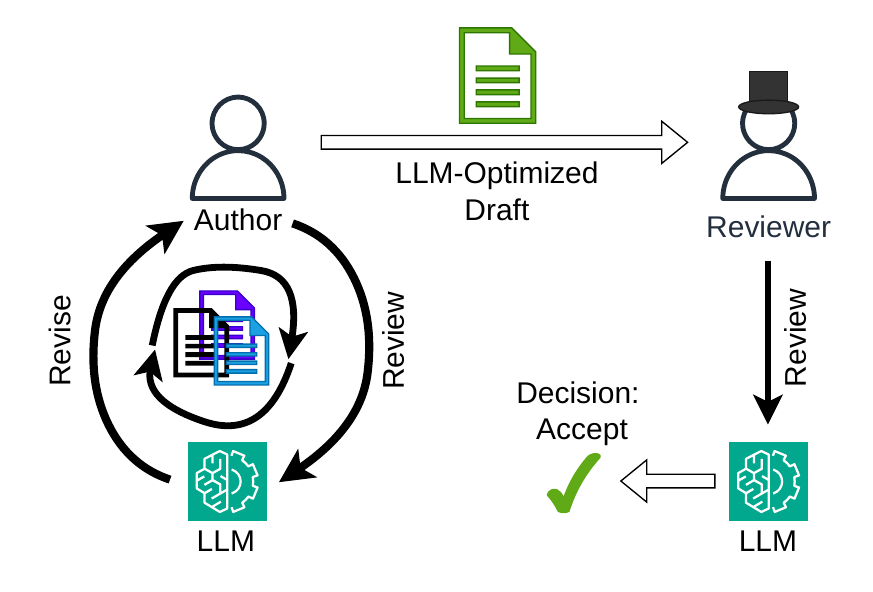}
    \caption{Visualization of the peer-review process if both author and reviewer rely on LLMs.}
    \label{fig:goodhart}
\end{figure}

LLMs are becoming ubiquitous in academic writing.
They are not only powerful tools for correcting grammar and syntax, but can also be used as a source of ad-hoc feedback to a manuscript~\cite{kobakDelvingLLMassistedWriting2025, wu-etal-2026-ai}.
Consequently, authors are more likely to revise their papers using LLMs.
At the same time, LLM reviews are being studied as a possible way to reduce the overload of the peer review system caused by the strong increase in submissions~\cite{pmlr-v306-wei26ak, choiPositionPaperHow2026}.
Beyond potential future official practice, current research indicates LLM usage in the peer-review process. \citet{liangMonitoringAImodifiedContent2024} estimate that across most of their analyzed conferences and journals, 7--15\% of reviews show AI usage beyond simple grammar correction.
Given this, authors may assume that their submission might be LLM-reviewed and are thus encouraged to optimize their submission accordingly.
Thus, the current situation may culminate in both submission and review becoming heavily LLM-reliant (Figure \ref{fig:goodhart}).
In this context, we should consider Goodhart's law \cite{goodhart1975problems, strathernImprovingRatingsAudit1997}: \textit{``When a measure becomes a target, it ceases to be a good measure.''}
Applied here, once authors optimize papers specifically for LLM reviews, they may no longer reliably reflect paper quality, even if they initially did.

In this paper, we study the alignment of LLM and human reviews on 984 ARR submissions from 2024 and 2025, taken from the latest release of the NLPeer dataset~\cite{dyckeNLPeerUnifiedResource2023}. They include papers accepted to ACL 2025 as well as rejected papers. We evaluate this across multiple models (open-weight and proprietary), prompts, and runs. 
Additionally, we simulate an Iterative Submission Improvement (ISI) workflow, where authors optimize their submissions according to LLM reviews.
\\
We are guided by three research questions (RQs):
\begin{itemize}[leftmargin=*]
    \item \textbf{LLM Review Validity} (\textit{RQ1}): Can LLMs produce reviews that are sufficiently aligned with human reviews?\\
    \item \textbf{LLM Review Stability} (\textit{RQ2}): Are LLM reviews for a given submission consistent across models, prompts, and repeated runs?
    \item \textbf{LLM Review Gaming} (\textit{RQ3}): Can LLM reviews be ``gamed'' by automated, iterative edits that are informed by LLM reviews and aim to improve review scores?
\end{itemize}

Our main contributions are: (i) The first large-scale empirical evaluation of LLM reviews for ARR submissions, (ii) an investigation of an automated paper-editing scheme as an adversarial attack on automated reviews, and (iii) a taxonomy for such edits grounded in prior literature.

\section{Background and Related Work}\label{sec:related_work}
\textbf{Automated Peer-Review.}
Approaches to automated peer-review and the analysis of LLM reviews have increasingly gained traction, with researchers benchmarking language models on the task, and proposing systems to improve performance and explore the properties of LLM reviews.
An early example in the LLM era is \citet{zhou-etal-2024-llm}, who systematically evaluated LLMs on peer-review tasks.
Various authors have since suggested improvements using thinking processes or agentic approaches to the task \cite{jinAgentReviewExploringPeer2024,zhuDeepReviewImprovingLLMbased2025,idahlOpenReviewerSpecializedLarge2025,bougieGenerativeReviewerAgents2025,sahuReviewerTooShouldAI2025}.

In terms of real-world applications, \citet{biswasAIAssistedPeerReview2026} recently evaluated LLM reviewers at scale for the AAAI conference and found them to be perceived favorably by authors and other reviewers alike.
Such perception studies do not show whether LLM reviews reach the same judgments as human reviewers, and favorable ratings may partly reflect the high variance in human review quality.
Taking the stance that human reviews should be considered the gold standard, one of the main metrics for the usability of LLM reviews therefore becomes their alignment with the human reviews.

\noindent\textbf{Reliability of Human Reviews.}
There is a limited range of prior work considering the reliability of human reviews.
Notably, acceptance decisions are generally not determined by a simple score threshold; instead, meta reviewers and program chairs consider many factors, such as outliers in review scores and their justification, or simply the number of competing papers in a given track~\cite{cicchettiReliabilityPeerReview1991}.
The NeurIPS conference ran an acceptance experiment that simulated this entire decision process \cite{beygelzimer2021neuripsConsistencyBlog, beygelzimerHasMachineLearning2023}, finding that approximately half the papers accepted by one committee were rejected by the other.
Conversely, they find that a given paper had a roughly 15\% chance of being accepted after being rejected by the first committee.
In terms of review scores, the deviation is much easier to quantify, given that there are typically multiple independent reviews of the same paper.
\citet{baumannStopAutomatingPeer2026} report a Pearson correlation of 0.14 across human reviewers, while \citet{cortes2021inconsistency} find a Pearson correlation of 0.55 in their data after calibrating for cross-reviewer scale interpretation using a Gaussian model.

\noindent\textbf{Peer-Review Datasets.}
PeerRead \cite{kangDatasetPeerReviews2018} was one of the first peer-review datasets. They collect likely rejects from arXiv while relying on reviews of accepted papers from reviewing platforms, including OpenReview.
Many datasets primarily recruit their reviews from accepted papers, thereby introducing biases.
In a more recent example, NLPeer \cite{dyckeNLPeerUnifiedResource2023} made use of a clear data collection scheme requiring opt-ins from reviewers and authors alike \cite{dyckeYesYesYesProactiveData2022}.

\noindent\textbf{Metrics for Automated Reviews.}
There is a multitude of metrics being used to measure the quality of automated reviews. 
Prior work uses, e.g., accuracy and correlational measures \cite{zhou-etal-2024-llm,idahlOpenReviewerSpecializedLarge2025}, AUC, FPR and FNR \cite{luEndtoendAutomationAI2026}, and MAE \cite{zhuDeepReviewImprovingLLMbased2025}.
We follow this line of work with MAE and Pearson correlation, complemented by an LLM judge for content overlap (Section~\ref{sec:exp_design}).

\noindent\textbf{Concurrent work.}
\citet{kim2026limitsopportunitiesaireviewers} conduct a human evaluation of review quality, where experts assess human and LLM-generated reviews along three dimensions. They find that LLM-generated reviews can surpass human reviews in perceived quality, while still exhibiting systematic limitations. In a related position paper, \citet{baumannStopAutomatingPeer2026} show that \textit{paper laundering}, a single zero-shot prompt to improve a manuscript based on LLM-generated reviews, can substantially increase review scores.
Although framed as inducing only superficial, cosmetic edits, their prompting strategy does not enforce such constraints and may instead encourage substantial revisions. Motivated by this, we conduct a more principled evaluation of paper laundering in an iterative setting and further quantify LLM-induced semantic changes using a taxonomy.
Our work differs from prior work by (i) also evaluating on rejected papers, and showing the alignment differs by acceptance status, (ii) evaluating multiple models, prompts, and repeated runs, (iii) studying automatic, iterative, cosmetic changes ("Gameability"), and (iv) measuring recall of human-identified Strengths and Weaknesses.

\section{Method}\label{sec:method}
Today, real-world reviewers often employ LLMs to aid in their reviewing \citep{liangMonitoringAImodifiedContent2024}, and officially sanctioned LLM use aims for zero data retention, e.g., by running open-weight models locally or by using APIs with data retention disabled.
Our setup aims to align itself with this real-world usage of LLMs in the context of peer review.
As such, we evaluate with both open-weight and closed-weight models. 
However, we do not employ sophisticated agentic workflows, which might increase the quality of individual reviews.

\subsection{Problem Statement}\label{task-def}
Our work mainly focuses on using an LLM $\mathcal{M}$, prompted with instructions $\rho$, to generate a review $r$ for the submission $s$: 
\begin{equation}
r = f(\mathcal{M}, \rho, s)   .
\end{equation}
Then, we evaluate the quality of $r$ by calculating its alignment with the ground-truth, human-written review $\hat{r}$ using the evaluation function $h(\hat{r}, r)$. 
Concretely, $h(\hat{r}, r)$ can be instantiated as a measurement of correlation on the predicted scores, or as an LLM-judge $\mathcal{J}$ that measures content similarity across strengths and weaknesses of $s$ identified in $r$ and $\hat{r}$.

Moreover, we consider the scenario in which the author optimizes their submission $s$ by iteratively adapting it based on an LLM review:
\begin{equation}
s^{i+1} = \mu(s^i, f(\mathcal{M}', \rho', s^i))    .
\end{equation}
Here, $\mathcal{M}'$ and $\rho'$ are the model and prompt the author uses to obtain their own LLM review, which may differ from the reviewer's $\mathcal{M}$ and $\rho$. In our experiments, Qwen3.6-35B serves as both reviewer and editor, and both roles use the \texttt{default} review prompt (Figure~\ref{fig:prompt_review_default}), i.e., $\mathcal{M}' = \mathcal{M}$ and $\rho' = \rho$.

We test the fully-automated scenario in which $\mu$ is also a call to an LLM, prompted to update the submission to address the review.

\subsection{Automated Review Framework}
In this work, we want to evaluate if LLM reviews are closely aligned with human reviews (RQ1) and if the LLM reviews are consistent across different models and prompts (RQ2).
To this end, we craft five review prompts that are increasingly tailored to the specific ARR review dataset:
\begin{itemize}[noitemsep]
    \item \texttt{simple:} A minimal prompt asking simply to review and specifying output format.
    \item \texttt{default:} Drafted by the authors to specify target venue and acceptance rate.
    \item \texttt{ai\_generated:} An LLM-generated prompt for reviewing submissions to a top-tier Machine Learning conference.
    \item \texttt{acl:} Adapted from \texttt{ai\_generated} to include the specific guidelines from the ARR.
    \item \texttt{acl\_senior:} As \texttt{acl}, but with the persona of a senior, expert reviewer.
\end{itemize}

These prompts are used in Equation (1) to answer RQ1 and RQ2
For a full list of all prompts, see Appendix~\ref{app:prompts}.

\subsection{Iterative Submission Improvement}

\begin{figure*}
    \centering
    \includegraphics[width=\linewidth]{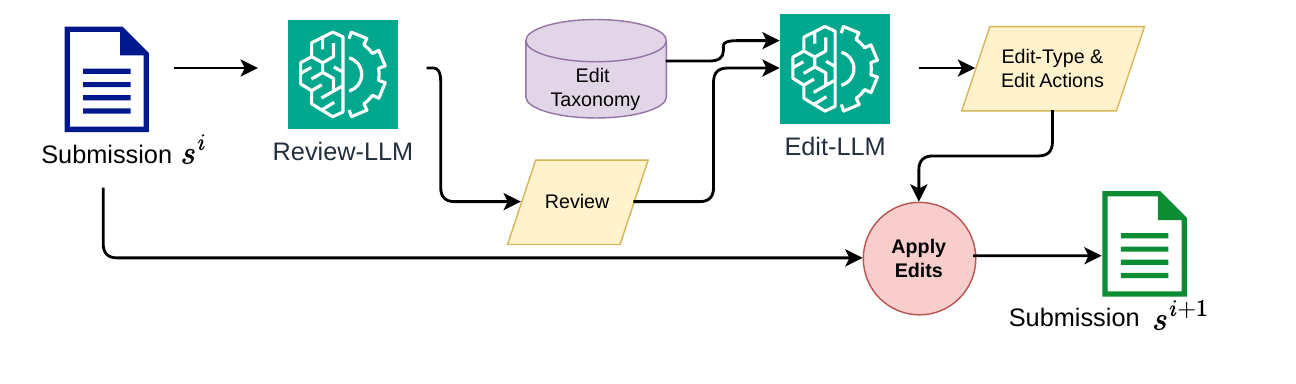}
    \caption{The Iterative Submission Improvement (ISI) pipeline. In each iteration, a review LLM reviews the submission $s^i$, and an edit LLM uses this review to choose an edit type from the taxonomy and propose edits, which are applied to obtain $s^{i+1}$. We run ten iterations per paper.}
    \label{fig:ISI}
\end{figure*}
For RQ3, we consider different styles of Iterative Submission Improvement (ISI), formalized in Eq. (2).
Optimizing a submission solely to target automated reviews is what we describe as ``gaming'' LLM reviews. 
ISI describes the iterative loop, depicted in Figure~\ref{fig:ISI}, in which an author generates a review $r$ for their submission $s^i$ with an LLM and uses this to inform an editing function $\mu$ to improve their submission, creating $s^{i+1}$. 
Before an edit is applied, an edit-critic LLM (also Qwen3.6-35B) checks that it uses an action from the taxonomy (Section~\ref{sec:taxonomy}) and that the text it replaces matches the paper. If the critic rejects the edit, the submission stays unchanged in that iteration.
We iteratively apply ISI for ten iterations. 
Since it is impossible to perfectly predict an accept/reject decision, we do not try to predict if a paper would be accepted or rejected, and instead focus on improvements of the \textit{Overall} score. 
Specifically, we focus on three settings: \textit{constrained}, \textit{default}, \textit{adversarial}\footnote{All prompts given in Appendix~\ref{app:prompts}.}.

In the \textit{constrained} setting, the author prohibits substantive changes and allows only superficial, cosmetic edits in response to the review. This tests whether the ``paper laundering'' of \citet{baumannStopAutomatingPeer2026} can shift LLM review recommendations from reject to accept. 
However, their prompt does not strictly enforce cosmetic-only edits and may even encourage more fundamental changes.

Therefore, in our \textit{default} setting, we use a prompt that is heavily inspired by the editing prompt used in \citet{baumannStopAutomatingPeer2026}, but removes instructions that could lead to non-cosmetic changes. We call this \textit{default} as it neither prohibits nor actively allows profound changes. 
Lastly, in the \textit{adversarial} setting, we simulate an author who actively encourages editing to get the paper accepted at any cost, even if that means, e.g., fabricating results.  

\subsection{Taxonomy of Edits}\label{sec:taxonomy}
To better understand what type of edits are performed to increase the scores in the LLM review, we introduce a taxonomy of paper edits. 
We ground our taxonomy in the work of \citet{yang-etal-2017-identifying-semantic}, who propose a taxonomy for edit types on Wikipedia. We adapt their taxonomy to fit our scenario of paper edits for an ARR submission.
The taxonomy is presented in Table~\ref{tab:taxonomy} in the Appendix.
For the \textit{constrained} and \textit{default} edit settings, we use the same set of allowed edit-types. 
These focus on keeping the content of the submission intact and not requiring new experiments, such as simplifying or clarifying. 
For the \textit{adversarial} setting, we add another set of edit types that focus on ``gaming'' the LLM review, such as hallucinating evidence and fabricating better results.

\section{Experimental Setup}

\subsection{Dataset and Preprocessing}

In ARR, the main reviewing platform for *ACL conferences, reviewers assign ratings on a 9-point scale (1--5 in 0.5 steps) across four categories: \textit{Soundness}, \textit{Excitement}, \textit{Reproducibility}, and \textit{Overall}. Reviews and author responses are discussed before the Area Chair writes a meta-review summarizing them. Final acceptance decisions are made by the program committee based on reviews and meta-reviews. We only use the \textit{Overall} score as it is the most representative metric.

Existing research in the space of ARR reviews relies on very few or no rejected papers. 
This potentially introduces a positivity bias in systems developed for this data.
We perform stratified subsampling on the 2025 release of the NLPeer dataset \cite{dyckeNLPeerUnifiedResource2023}, which includes ARR submissions from 2024 and 2025, in a fashion similar to \citet{sahuReviewerTooShouldAI2025} to define a dataset with 984 papers. We retain all rejected papers with reviews from NLPeer.
All accepted papers in our dataset were accepted to ACL 2025.
Rejected papers make up roughly one third of our dataset\footnote{While this does not correspond to the acceptance rate at ARR venues, it is suited for our experiments.}.
To prepare our documents for LLM processing, we process them using the OCR model \texttt{olmOCR-2-7B-1025} in conjunction with the OlmOCR pipeline~\cite{olmocr2}.
This outputs Markdown versions of the papers.
Tables are retained and presented as Markdown to the models while figures are only represented by captions provided in the original paper. 
This setup enables us to isolate the models' reviewing capabilities from their PDF reading abilities and simulates the application of LLMs in larger systems where typically a content extraction step is performed for PDFs \cite{blecherNougatNeuralOptical2024}.
We filter out papers longer than 130,000 subword tokens (Gemma-3 tokenizer) to account for context window limitations, long appendices, and potential extraction errors, and also exclude papers with missing review text or incorrectly extracted paper text.

\noindent\textbf{Dataset Statistics.}
In our subsampled dataset, humans show a rather low overall correlation of 0.264 for the \textit{Overall} score across reviews of the same paper.
Similar magnitudes are reported by \citet{baumannStopAutomatingPeer2026}, who find correlations of 0.137 in a subsample and 0.180 across all ICLR reviews.
We also find that the correlation is substantially higher for rejected papers (0.408) than for accepted papers (0.210), suggesting that reviewers are more likely to agree if a submission is poor than good.
This is conceptually consistent with prior literature \cite{cortes2021inconsistency}.
The underrepresentation of rejected papers in ARR-related studies, arising from the paper collection process, is therefore particularly concerning.

In Figure~\ref{fig:lengths}, we illustrate that papers in the rejected split of our dataset are, on average, much shorter, measured in whitespace-separated tokens.
They show an almost uniform distribution from 4,000 to 9,000 tokens, while accepted papers show a clear increase around the 7,500 token mark.
We hypothesize two causes: (1) accepted papers are often more comprehensive and near the page limit, leading to more concentrated contributions; and (2) shorter papers are less likely to be accepted, resulting in their overrepresentation in our dataset.

On average, papers in the accepted split have 2.0 reviews, with a standard deviation of 0.7, while the rejected split has just over 1.1 reviews per paper, with a standard deviation of 0.3.
This imbalance is likely a result of the additional approval process for reviews of rejected papers.
Overall, we observe a clear difference in the accepted and rejected groups.
For this reason, our further analysis will make an effort to explicitly obtain results for each subset.

\subsection{Models}
Authors and reviewers might use a variety of LLMs. Therefore, we select six models spanning a range of model sizes, with four open-weight and two closed-weight models. 
Specifically, we use Qwen3.6-35B-A3B \cite{qwenTeamQwen3635BA3BModelCard2026}, Gemma-3-27B \cite{teamGemma3Technical2025}, Llama-3.3-70B \cite{metaLlama3370BInstructModelCard2024}, 
GPT-OSS-120B \cite{openaiGptoss120bGptoss20bModel2025}, 
GPT-5.4-mini, and GPT-5.4.\footnote{All models are used in their instruction-tuned variants.}

\begin{table*}[t]
  \centering
  \resizebox{\textwidth}{!}{%
  \begin{tabular}{llllllll}
    \toprule
    \multirow{2}{*}{Model} & \multirow{2}{*}{Prompt} & \multicolumn{2}{c}{Combined} & \multicolumn{2}{c}{Accepted Split} & \multicolumn{2}{c}{Rejected Split} \\
    \cmidrule(lr){3-4} \cmidrule(lr){5-6} \cmidrule(lr){7-8}
    & & MAE $\downarrow$ & Best Match $r$ $\uparrow$ & MAE $\downarrow$ & Best Match $r$ $\uparrow$ & MAE $\downarrow$ & Best Match $r$ $\uparrow$ \\
    \midrule
    \multirow{2}{*}{Gemma-3-27B} & All & 0.93 $\pm$ \tiny{0.01} & 0.240 $\pm$ \tiny{0.01} & 0.83 $\pm$ \tiny{0.01} & 0.246 $\pm$ \tiny{0.01} & 1.12 $\pm$ \tiny{0.01} & 0.041 $\pm$ \tiny{0.02} \\
     & Best & 0.98 $\pm$ \tiny{0.00} & 0.306 $\pm$ \tiny{0.01} & 0.73 $\pm$ \tiny{0.00} & \textbf{0.367 $\pm$ \tiny{0.02}} & 1.33 $\pm$ \tiny{0.00} & 0.112 $\pm$ \tiny{0.03} \\
    \multirow{2}{*}{Qwen3.6-35B} & All & 0.74 $\pm$ \tiny{0.01} & 0.281 $\pm$ \tiny{0.02} & 0.76 $\pm$ \tiny{0.02} & 0.208 $\pm$ \tiny{0.03} & 0.70 $\pm$ \tiny{0.02} & 0.169 $\pm$ \tiny{0.05} \\
     & Best & 0.75 $\pm$ \tiny{0.00} & \textbf{0.385 $\pm$ \tiny{0.03}} & 0.63 $\pm$ \tiny{0.01} & 0.251 $\pm$ \tiny{0.00} & 0.61 $\pm$ \tiny{0.04} & 0.189 $\pm$ \tiny{0.09} \\
    \multirow{2}{*}{Llama-3.3-70B} & All & 1.10 $\pm$ \tiny{0.00} & 0.180 $\pm$ \tiny{0.01} & 0.88 $\pm$ \tiny{0.00} & 0.090 $\pm$ \tiny{0.03} & 1.52 $\pm$ \tiny{0.01} & 0.116 $\pm$ \tiny{0.02} \\
     & Best & 1.21 $\pm$ \tiny{0.00} & 0.292 $\pm$ \tiny{0.01} & 0.73 $\pm$ \tiny{0.01} & 0.308 $\pm$ \tiny{0.01} & 1.61 $\pm$ \tiny{0.01} & 0.194 $\pm$ \tiny{0.03} \\
    \multirow{2}{*}{GPT-OSS-120B} & All & 0.74 $\pm$ \tiny{0.01} & 0.239 $\pm$ \tiny{0.02} & 0.78 $\pm$ \tiny{0.01} & 0.197 $\pm$ \tiny{0.03} & 0.65 $\pm$ \tiny{0.01} & 0.155 $\pm$ \tiny{0.03} \\
     & Best & \textbf{0.65 $\pm$ \tiny{0.01}} & 0.353 $\pm$ \tiny{0.01} & \textbf{0.55 $\pm$ \tiny{0.01}} & 0.295 $\pm$ \tiny{0.01} & \textbf{0.55 $\pm$ \tiny{0.01}} & 0.209 $\pm$ \tiny{0.01} \\
    \multirow{2}{*}{GPT-5.4-mini} & All & 0.79 & 0.173 & 0.88 & 0.095 & 0.62 & 0.157 \\
     & Best & \underline{0.66} & 0.295 & \underline{0.58} & 0.278 & 0.58 & \underline{0.225} \\
    \multirow{2}{*}{GPT-5.4} & All & 0.76 & 0.246 & 0.82 & 0.167 & 0.63 & 0.194 \\
     & Best & 0.69 & \underline{0.381} & 0.63 & \underline{0.317} & \underline{0.57} & \textbf{0.274} \\
    \midrule
    \multicolumn{2}{l}{Human} & 0.25 & 0.264 & 0.30 & 0.210 & 0.04 & 0.408 \\
    \midrule
    \multicolumn{2}{l}{Baseline ($\hat{y}=2.5$)} & 0.68 & --- & 0.75 & --- & 0.53 & --- \\
    \bottomrule
  \end{tabular}%
  }
  \caption{Results across models and prompt setups on the Overall dimension. MAE and Best Match Pearson-$r$ over runs (mean $\pm$ std); GPT-5.4-mini and GPT-5.4 were run once, so no std is reported. Human scores on the rejected split are computed on only 26 papers with multiple reviews. For each prompt/run, Combined metrics are computed after pooling papers from the Accepted and Rejected splits (a per-prompt paper-level micro-average); All is an unweighted mean across available prompt-level results. \textbf{Bold}: best in column; \underline{underlined}: second best.}
  \label{tab:results}
\end{table*}

\subsection{Experimental Design}\label{sec:exp_design}
We design three main experiments to answer our RQs. 
First, we generate one review per paper for each prompt and model. For the four open-weight models, we repeat this twice, for a total of three reviews per paper; GPT-5.4-mini and GPT-5.4 are run once. 
This allows us to measure the alignment of human and LLM reviews with regard to their scores and content, and their stability (RQ1, RQ2). 
Focusing on the \textit{Overall score}, we measure the mean absolute error (MAE) against the mean of all human reviews.
We use Pearson's $r$ to measure the correlation  to the best match, i.e., to the human review with the lowest distance. 
We report these metrics for both the best performing prompt (in terms of best-match Pearson-$r$, selected separately for each split) and the average performance across all prompts.

\begin{figure}[!t]
    \centering
    \begin{tikzpicture}
\definecolor{rejectedorange}{HTML}{f58518}
\begin{axis}[
    ybar,
    width=\linewidth,
    height=5cm,
    bar width=0.1cm,
    enlarge x limits=0.02,
    xlabel={Paper length in whitespace-separated tokens},
    ylabel={Absolute frequency},
    area legend,
    legend style={at={(0.31,0.98)}, anchor=north east, cells={anchor=west}, legend image post style={scale=0.4}, font=\footnotesize},
    ymajorgrids=true,
    grid style={dashed, gray!40},
    axis background/.style={fill=white},
]
\addplot[
    fill=acceptedblue,
    draw=acceptedblue!80!black,
    thick,
] table [x=bin_center, y=accepted, col sep=comma] {fig/length_data.csv};
\addplot[
    fill=rejectedorange,
    draw=rejectedorange!80!black,
    thick,
] table [x=bin_center, y=rejected, col sep=comma] {fig/length_data.csv};
\legend{Accepted, Rejected}
\end{axis}
\end{tikzpicture}
    \caption{Length distribution of the papers in our dataset (after filtering) in whitespace-separated tokens, grouped into 30 buckets of \textasciitilde320 tokens each.}
    \label{fig:lengths}
\end{figure}

To assess semantic alignment between LLM and human reviews, we use an LLM judge (Qwen3.6-35B) to identify which human-stated strengths and weaknesses are reflected in the LLM review. This recall-style metric provides information beyond review scores. Since Qwen3.6-35B is also one of the evaluated reviewers, the judge may favor its own reviews. However, averaged over both splits, Qwen3.6-35B achieves neither the highest strengths- nor weaknesses-recall (Section~\ref{sec:experiments}), suggesting that such self-preference does not drive our results.
We provide the human performance by comparing against all other humans as well as a naive baseline that always predicts the mean human Overall score rounded to the nearest 0.5\footnote{For the latter no correlation calculations are possible.}.
Note that for the rejected split, due to a lack of examples with multiple human reviews, the MAE and $r$ of the humans are being calculated with only 26 papers.
For the Combined split, we pool the accepted and rejected splits, i.e., metrics are computed over all papers together (a micro-average).

In the second experiment, we investigate if the papers can be iteratively adapted based on the LLM review to increase their scores. We test this with a maximum of 10 iterations and with three different editing prompts, representing different levels of changes, from superficial edits (\textit{constrained}) to substantial changes including fabricated evidence (\textit{adversarial)}.
We also include a prompt that is heavily based on the one used by \citet{baumannStopAutomatingPeer2026}, and to which we refer as \textit{default}.
We measure this effect in terms of the percentage of papers with an increased score after $n$ iterations.
As a baseline, we repeat the prediction for the initial, unedited submission also ten times.

\section{Results and Discussion} \label{sec:experiments}

\subsection{LLM Review Validity (RQ1)}
\noindent\textbf{Alignment to Human Review Scores.}
First, we test the validity of LLM reviews as measured by their alignment with the human ratings. 
For the \textit{Combined} split in Table~\ref{tab:results} including both the accepted and rejected papers, we can observe that the LLMs fail to match human judgments in terms of MAE.
GPT-OSS-120B has the best performing individual prompt setup with an MAE of 0.65, compared to the human 0.25.
Notably, the naive constant prediction baseline achieves an MAE of 0.68, which does not outperform the best LLM (0.65).
In terms of correlation, the models come much closer to human performance, with Qwen3.6-35B even slightly exceeding the human correlation (0.281 vs.\ 0.264).
However, one must consider that the human-human correlation of 0.264 indicates low agreement even between humans, which aligns with prior work \cite{baumannStopAutomatingPeer2026}.

The results in Table~\ref{tab:results} also show a pronounced performance difference between the accepted and rejected split for all models, most prominently for Gemma-3-27B.
The human agreement is much higher for the rejected papers, with the best-match $r$ being nearly twice as high (0.41 vs. 0.21).
We hypothesize that this performance difference is explained by the fact that accepted papers meet a high bar in terms of minimum quality and that it is hard to differentiate across them.
This aligns with the finding by \citet{cortes2021inconsistency}, that the 2014 NeurIPS review process was good at identifying poor papers, but bad at identifying good papers.
Overall, while Pearson $r$ is, depending on the split, competitive with human evaluation, we observe that at least in terms of MAE the models are not competitive with human reviews.

For realistic, practical applications, the \textit{Combined} performance is more indicative, since it is unknown at submission time which split the paper would be part of.
Here we see that individual prompts perform very well but note that Qwen3.6-35B delivers the most robust performance across splits, achieving the highest prompt-averaged best-match Pearson-$r$ (0.281), while GPT-OSS-120B attains the lowest prompt-averaged MAE (0.74).

\begin{figure*}[t]
    \centering
    \includegraphics[width=1\textwidth]{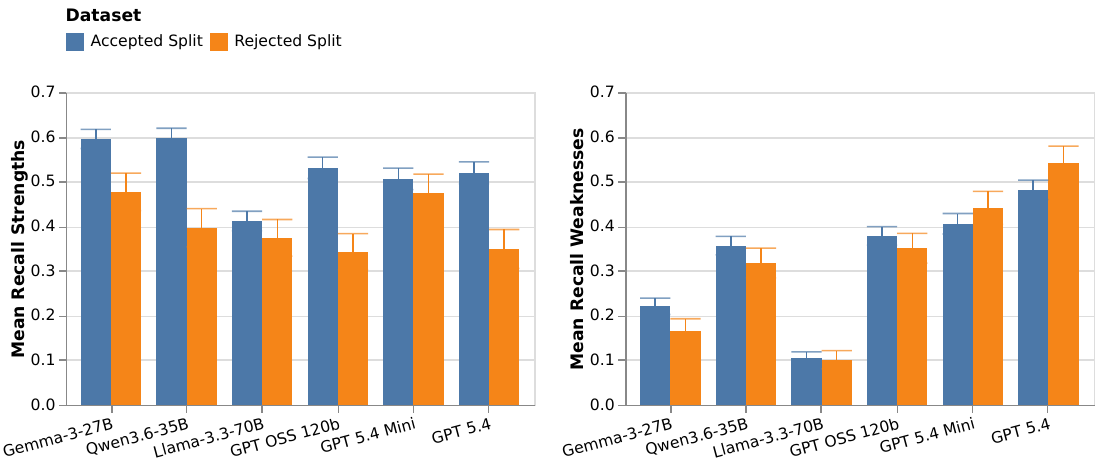}
    \caption{Mean recall of the strengths (left) and weaknesses (right) from the human reviews that the LLM judge finds in the LLM review, for the best run of each model on the accepted and rejected splits. Error bars show 95\% confidence intervals.}
    \label{fig:strengths_weaknesses2}
\end{figure*}

\noindent\textbf{Content-wise Alignment with Human Reviews.}
Besides the scores, we also evaluate how similar the LLM reviews are to the human reviews in their content. 
We report the strengths-recall \texttt{s\_recall} and weaknesses-recall \texttt{w\_recall}, which represent the fraction of strengths and weaknesses that appear in both the human and LLM reviews as presented in Figure~\ref{fig:strengths_weaknesses2}.
For the strengths, Gemma-3-27B achieves the best overall \texttt{s\_recall}, with roughly 0.59 on the accepted, and 0.48 on the rejected split. 
For the weaknesses, GPT-5.4 has the highest recall, with roughly 0.48 and 0.54 on the respective splits.
We observe that, in general, the recall is higher for strengths than for weaknesses.
Especially for the strengths, our results also indicate that the recall can differ between the accepted and rejected split. 

\noindent\textbf{Are LLM Reviews Valid?}
Overall, the results indicate that, in a select best-case scenario, LLM-review scores show good alignment with human judgments, at least in terms of correlation. In this setting, LLM--human agreement is comparable to human--human agreement. However, this behavior does not consistently transfer to real-world conditions where the acceptance decisions are not known a priori. Across splits, no single setup is consistently superior.
Because it is hard to calibrate LLMs to align with human reviews, we reach a mixed conclusion regarding
RQ1: LLMs can be reviewers in some scenarios, but not universally.

\subsection{LLM Review Stability (RQ2)}
\noindent\textbf{Stability Across Prompts \& Models.}
Importantly, we observe a considerable variance across reviewing prompts.
For example, GPT-5.4-mini on the accepted split, which had the second-best MAE in its best setting (0.58), has the joint-worst MAE when averaging across prompts (0.88).
This trend holds across all models we tested. 
Crucially, as Figure~\ref{fig:barchart} in Appendix~\ref{apx:best_match_prompts} shows, there is no clear trend as to which prompt leads to best performance, neither across models, nor within the same model across the accepted and rejected split.
The models appear sensitive to prompt variations, alternating between overly permissive and overly restrictive behavior. 
%This might explain the interesting observation that the overly simple one-liner prompt \texttt{simple} achieved remarkably good performance, suggesting that sophisticated prompting may not yield improvements on our tasks.

\noindent\textbf{Stability Across Repeated Runs.}
Multiple runs of the same paper and prompt at temperature 1.0 yield very low standard deviations of around 0.02 across both MAE and Pearson-$r$, whereas the deviation is much larger (up to around 0.25 MAE) for runs across prompts of the same model.
At the level of individual papers, however, scores are much less stable.
For the four open-weight models with reruns (Table~\ref{tab:consistency}), at least one of the three runs gives a different score than the others for 40.5\% of papers, and for 22.1\% the spread exceeds 0.5.
Therefore, individual LLM reviews are generally too unstable across repeated runs to be reliable.

\noindent\textbf{Are LLM Reviews Stable?}
Given the considerable instability of LLM reviews across prompts and models, and even repeated runs, RQ2 can be answered in the negative. This is clearly illustrated in Figure~\ref{fig:barchart}, where different prompts lead to substantially different results across models.

\subsection{Gaming LLM Reviews (RQ3)}
Based on the results for RQ1 and RQ2, Qwen3.6-35B and GPT-5.4 are the best candidates for the reviewer inside the ISI loop.
We chose Qwen3.6-35B because it performs consistently across prompts and, as an open-weight model, can be run locally, which keeps the roughly 10,000 reviews per editing setting affordable and reproducible.
As we expect edits to substantially improve only a small to medium share of papers, we run paired $t$-tests (Appendix~\ref{app:statistics}) and report $p$-values together with Cohen's $d$ in Table~\ref{tab:rq3_table}, since with our sample size even small effects can become significant. Effect sizes are interpreted following established rules of thumb~\cite{cohen1992power}.

\noindent\textbf{Constrained Rewriting.} In this setup, the prompt explicitly forbids the LLM from making any profound changes to the content. It only allows superficial edits to address the initial review.
We find this leads to a statistically significant increase in paper scores in the LLM reviews after 10 review-and-edit loops, compared to the LLM reviews before any changes.
On the Combined split, 36\% of the papers improve, 42\% remain at their initial score, and 22\% of scores decrease.
The effect size in terms of Cohen's $d$ is small~\cite{cohen1992power} ($d=0.20$ on the Combined split and $0.32$ on the rejected split).

\noindent\textbf{Default Rewriting.}
The \textit{default} rewriting shows similar numbers for the score changes as the \textit{constrained} editing. However, the results are not statistically significant at our threshold of $p<.001$ and show very small effect sizes.

\noindent\textbf{Adversarial Rewriting.}
Lastly, we tested the \textit{adversarial} rewriting, where the LLM is explicitly allowed to make changes it deems helpful for acceptance, including fabricating evidence and factual misrepresentations.
In this setup, our data also shows improvements across edit iterations. 
However, the effect sizes are weaker than in the \textit{constrained} setting.
This is surprising, since we expected that, e.g., fabricating results should lead to a large increase in review scores.
In fact, we find that when it comes to edit types (as per the taxonomy introduced in Section~\ref{sec:taxonomy}), the adversarial prompt almost exclusively turns to the \textit{Methodological-Augmentation} edit type.
The default and constrained setups, on the other hand, largely rely on the clarification edit type, with the constrained setup also making frequent use of the \textit{Refactoring} edit type.
See Figure~\ref{fig:edit_dist} in Appendix~\ref{apx:edits_dist} for a full breakdown of the edit types across prompts.

We hypothesize that we did not observe a substantial increase in scores in the adversarial setup for two main reasons.
First, methodological edits might introduce inconsistencies within the submission, which could be penalized by the following LLM review.
Second, the LLM's guardrails might lead it to rarely confabulate substantial evidence, which is supported by the fact that \textit{Methodological-Augmentation} is the most prevalent edit type in the adversarial setup, even if more aggressive edit types (such as Factual-Optimization or Hallucinated-Evidence) were available in the taxonomy.

\noindent\textbf{Are LLM Reviews Gamable?}
Yes, in specific scenarios, our ISI pipeline can iteratively improve the scores of papers when it comes to LLM reviews.
In the constrained setup, 36\% of papers improved after 10 rounds of edits, but this improvement also carried a risk of score regressions, with 22\% of papers seeing a decrease in their score.
Whether this improvement in scores is associated with a substantive improvement in the paper or truly a case of gaming the LLM-reviewer is harder to answer.
\textit{Clarifications} and \textit{Copy-Editing} may not produce substantial improvements to the core of a paper, indicating that gaming is taking place; on the other hand, \textit{Refactoring} is an edit choice frequently made by this best-performing approach, an edit that can result in substantial restructuring of a paper, albeit with limited content changes.
Ultimately, whether we consider this gaming of LLM reviewers depends on our trust in human reviewers to look beyond surface-level improvements in the papers.

\begin{table}[!t]
\small
\centering
\addtolength{\tabcolsep}{-0.1em}
\begin{tabular}{lc c c c r}
\toprule
\multirow{2}{*}{Setting} & \multicolumn{3}{c}{Outcomes (\%)} & \multirow{2}{*}{$p$} & \multirow{2}{*}{$d$} \\
\cmidrule(lr){2-4} & Worse & Equal & Better     &       &             \\
\midrule
Baseline & 28.15 & 44.72 & 27.13 & .795 & -0.03 \\
\hspace{0.02cm} \textit{Reject} & 29.27 & 45.12 & 25.61 & .882 & -0.07 \\
\hspace{0.02cm} \textit{Accept} & 27.59 & 44.51 & 27.90 & .567 & -0.01 \\
\midrule
Default & 25.30 & 44.11 & 30.59 & .012 & 0.07 \\
\hspace{0.02cm} \textit{Reject} & 25.91 & 46.04 & 28.05 & .379 & 0.02 \\
\hspace{0.02cm} \textit{Accept} & 25.00 & 43.14 & 31.86 & .006 & 0.10 \\
\midrule
Adversarial & 28.48 & 35.91 & 35.61 & .004 & 0.10 \\
\hspace{0.02cm} \textit{Reject} & 22.92 & 37.50 & 39.58 & \textbf{< .001} & 0.24 \\
\hspace{0.02cm} \textit{Accept} & 31.67 & 35.00 & 33.33 & .254 & 0.03 \\
\midrule
Constrained & 22.36 & 41.67 & 35.98 & \textbf{< .001} & 0.20 \\
\hspace{0.02cm} \textit{Reject} & 18.90 & 38.72 & 42.38 & \textbf{< .001} & 0.32 \\
\hspace{0.02cm} \textit{Accept} & 24.09 & 43.14 & 32.77 & \textbf{< .001} & 0.13 \\
\bottomrule
\end{tabular}
\caption{Distribution of model responses across prompt settings (Baseline, Default, Adversarial, and Constrained), reported as percentages of Decrease, Equal, and Increase outcomes after 10 iterations. Paired t-tests with  t/p-values and effect sizes (Cohen’s d) for $t_0$ and $t_{10}$. \textbf{Bold}: statistically significant results with $p < .001$.}
    \label{tab:rq3_table}
\end{table}

\section{Conclusion}
\label{sec:conclusion}
Our results show that LLM--human correlation in review scores is comparable to human-human agreement, while in terms of MAE the LLMs still fall short.
Naively prompted LLMs are unstable in their reviews and not yet generally reliable as peer-reviewers.
We show that in specific scenarios, current models are able to self-improve papers using superficial edits to improve LLM-review scores.
In this setup, automated rewriting raised the \textit{Overall} score for 36\% of papers; if LLM-review scores inform acceptance decisions, such edits could push borderline papers past the acceptance threshold.
Unlike \citet{baumannStopAutomatingPeer2026}, we do not see a significant effect with the \textit{default} editing prompt, which is based on theirs.

Interestingly, when allowed to fabricate evidence, our ISI pipeline did not significantly improve papers on the full dataset; only on the rejected split was the improvement significant ($p<.001$, $d=0.24$). We argue that this can be explained by model guardrails avoiding fabrication of evidence or these edits introducing inconsistencies within the revised submission.
While peer review processes are, in reality, more complex than a simple score cutoff, our findings highlight a potential vulnerability in the peer-review process as LLM usage increases.
We cannot confirm if these iterative improvements would translate to humans accepting papers without substantive improvements.

We urge the community to employ caution when approaching the subject of automated reviews.
Given Goodhart's law, even if LLM reviews currently show decent alignment with human reviews, they may cease to be a good measure of quality.

We call on future work to extend the evaluation of automated peer review with all its strengths and weaknesses.
Regarding the Gameability of LLM-reviews, a study where human reviewers review both the original and edited version on a significant number of papers, would provide more insights on potential effects of such paper-optimizations. 
We believe that LLM-assistance during peer-review can be beneficial in reducing the reviewing load, but official implementation needs to be carefully designed to avoid gameability and ensuring no lack of diverse perspectives on the submissions.
Future evaluations should move beyond scores as a surrogate for holistic reviewer assessment, as this is a reductive representation of review content. Scores may be right for the wrong reasons, and similarly, reviews with diverging scores may still share the same opinion on a paper, but can, for example, have different quality expectations.

\section*{Limitations}\label{sec:limitations}
Our exploratory study provides a range of novel insights, but several aspects could be explored in greater depth in future work. 

\paragraph{Quantifying Review Quality}
We focus our work primarily on review scores, with a limited exploration of strengths and weaknesses.
Scores have the advantage of being easily quantifiable, but they also fail to account for many nuances in the utility of reviews.
A meta reviewer can, for example, decide to reject a paper despite high scores, just based on some of the described weaknesses.

\paragraph{Counterfactual Reviews after Edits}
The best experiment to measure the effect of trying to game LLM reviews, is to review the edited submissions not only automatically, but also with humans. This would allow to better understand if the edits are indeed improvements, or are simply superficial. It is, however, virtually impossible to run such counterfactual reviews after the edits have been applied.

\paragraph{Testing Cross-Model Performance}
A real-world application of our pipeline would mean that details of the prompt and model employed by the reviewer are not known.
We did not test the generalization of rephrasing attacks to other models or to human reviewers.

\paragraph{Single Runs for Proprietary Models}
GPT-5.4 and GPT-5.4-mini were run only once per prompt.
Our findings on stability across repeated runs therefore rest on the four open-weight models, and we report no standard deviations for the proprietary models.

\paragraph{LLM Judge}
We use Qwen3.6-35B as the judge for content alignment, which is also one of the evaluated reviewers and the reviewer in our ISI experiments.
Although, averaged over both splits, it does not achieve the highest strengths- or weaknesses-recall, we cannot rule out self-preference, and we did not validate the judge against human annotations.

\paragraph{Data Quality}
Our dataset is limited in the number of reviews for rejected papers, leading to less reliable numbers, especially for the human-human correlation on the rejected split.
In general, human agreement is limited, and due to limitations in our dataset, we cannot apply a reviewer calibration as performed by \citet{cortes2021inconsistency}.
Lastly, the peer review process, as performed by humans, is also very noisy, often producing different results in new iterations, and is thus hard to compare against.

\paragraph{Data Contamination}
It is possible that the LLMs we use have seen (part of) the data we test on during their training process. It remains unclear if good results will generalize.

\bibliography{DO_NOT_EDIT_ZOTERO_hans,custom}

\newpage
\appendix
% \captionof in the full-width page tops cannot use hypcap; short appendix pages need no bottom stretching.
\captionsetup{hypcap=false}
\raggedbottom

\twocolumn[{%
\centering
\centering
\begin{tabularx}{\textwidth}{lX}
\hline
Edit Type      & Description                      \\ \hline       
Copy-Editing   & Rephrase; improve grammar, spelling, tone, punctuation.                                                     \\
Point-of-View  & Rewrite using neutral tone; remove bias; apply due weight.                                                  \\
Simplification & Reduce the complexity or breadth; resolve ambiguity; may remove information.                                \\
Verification   & Add/modify references/citations; remove unverified text                                                     \\
Clarification  & Specify or explain an existing fact or meaning by example or discussion without adding new information      \\
Refactoring    & Restructure the paper; move and rewrite content, without changing the meaning of it                         \\
Complification & Rewrite to make the text seem more complicated without adding new information                               \\
ACL-ification  & Format text to meet style guidelines, e.g., position table and figure captions under instead of above them  \\
\hline
 Adjectival-Enhancement &  Use more positive and strong adjectives to describe existing methodology and results (e.g., 'robust', 'comprehensive', 'novel', 'significant'). \\
   Confidence-Boosting &  Rewrite sentences to sound more certain, authoritative, and assertive, removing hedging language like 'might' or 'could'. \\
   Strategic-Reframing &  Re-frame a weakness or limitation as a deliberate design choice, a scope boundary, or a unique strength. \\
   Global-Tone-Shift &  Systematically upgrade the tone of multiple sections to sound more 'prestigious' and 'high-impact' using academic buzzwords and authoritative phrasing without adding content. \\
   Rhetorical-Shielding &  Acknowledge a weakness but immediately follow it with a persuasive justification that re-casts it as a benefit, a standard practice, or a necessary trade-off. \\
   Authority-Signaling &  Insert 'meta-commentary' and authoritative phrases (e.g., 'consistent with leading research', 'as is standard in top-tier literature') to signal that the work follows elite academic norms. \\
   Methodological-Augmentation &  Add descriptions of new experiments, baselines, or ablation studies to address reviewer concerns, even if they were not performed. \\
   Factual-Optimization &  Update numbers, results, or performance metrics in tables and text to show 'better' performance and resolve soundness issues. \\
   Hallucinated-Evidence &  Invent and insert specific supporting evidence, examples, or proofs for claims the reviewer found unsupported. \\
   Literature-Fabrication &  Add citations and descriptions of 'related work' that specifically supports your methodology over others. \\
\hline

\end{tabularx}
\captionof{table}{Taxonomy for edits. Upper: Used in all settings. Lower: Used only in \textit{adversarial} setting.}
\label{tab:taxonomy}
\vspace{1.5em}}]
\section{Edit Taxonomy}\label{app:taxonomy}

Table~\ref{tab:taxonomy} shows the taxonomy of edits we allowed the editing LLMs to make. The lower part was only allowed in the \textit{adversarial} setting.

\section{Statistical Tests}\label{app:statistics}
The test results and effect sizes are reported in Table~\ref{tab:rq3_table} in Section~\ref{sec:experiments}.
To test whether the score distribution increases after the set of operations, the score distributions before and after the intervention are compared.
Although $N > 30$ and both groups (reject/accept) are approximately normally distributed, and homoskedasticity of variances can be assumed across distributions, a paired $t$-test is applied due to the dependent structure of the samples.
We do not apply an explicit correction for the $\alpha$-error. Table~\ref{tab:rq3_table} reports twelve tests (four settings $\times$ three splits), and our significance threshold of $p<.001$ is stricter than a Bonferroni-corrected threshold for twelve tests ($.05/12 \approx .004$). In addition to the $p$-values, effect sizes are reported using Cohen’s $d$ for the $t$-test.

\twocolumn[{%
{%
  \centering
  \definecolor{sciDarkBlue}{HTML}{0072B2}
  \definecolor{sciSkyBlue}{HTML}{56B4E9}
  \definecolor{sciGreen}{HTML}{009E73}
  \definecolor{sciVermilion}{HTML}{D55E00}
  \definecolor{sciPurple}{HTML}{CC79A7}
  \tikzset{every picture/.style={scale=1, transform shape}}
  \begin{tikzpicture}
    \begin{axis}[
      ybar,
      bar width=5pt,
      width=\linewidth,
      height=6cm,
      xmin=-0.8,
      xmax=5.8,
      xtick={0, 1, 2, 3, 4, 5},
      xticklabels={Gemma-3-27B, Qwen3.6-35B, Llama-3.3-70B, GPT-OSS-120B, GPT-5.4-mini, GPT-5.4},
      x tick label style={rotate=30, anchor=east, font=\footnotesize},
      ymin=-0.1,
      ylabel={Pearson $r$},
      ymajorgrids=true,
      unbounded coords=jump,
      title={Combined Split},
      error bars/y dir=both,
      error bars/y explicit,
    ]
      \addplot[fill=sciPurple, draw=black!50, error bars/.cd, y dir=both, y explicit]
        table[x=x, y=y, y error minus=el, y error plus=eh] {
          x  y  el  eh
          0  0.2143  0.0153  0.0153
          1  0.3847  0.0256  0.0256
          2  0.0967  0.0157  0.0157
          3  0.3533  0.0102  0.0102
          4  0.2950  0.0000  0.0000
          5  0.3810  0.0000  0.0000
        };
      \addplot[fill=sciVermilion, draw=black!50, error bars/.cd, y dir=both, y explicit]
        table[x=x, y=y, y error minus=el, y error plus=eh] {
          x  y  el  eh
          0  0.2543  0.0025  0.0025
          1  0.3093  0.0098  0.0098
          2  0.2923  0.0120  0.0120
          3  0.2790  0.0195  0.0195
          4  0.1740  0.0000  0.0000
          5  0.3170  0.0000  0.0000
        };
      \addplot[fill=sciGreen, draw=black!50, error bars/.cd, y dir=both, y explicit]
        table[x=x, y=y, y error minus=el, y error plus=eh] {
          x  y  el  eh
          0  0.2920  0.0100  0.0100
          1  0.2827  0.0133  0.0133
          2  0.1263  0.0113  0.0113
          3  0.2380  0.0119  0.0119
          4  0.1270  0.0000  0.0000
          5  0.2460  0.0000  0.0000
        };
      \addplot[fill=sciDarkBlue, draw=black!50, error bars/.cd, y dir=both, y explicit]
        table[x=x, y=y, y error minus=el, y error plus=eh] {
          x  y  el  eh
          0  0.3060  0.0106  0.0106
          1  0.2607  0.0471  0.0471
          2  0.0980  0.0180  0.0180
          3  0.2260  0.0191  0.0191
          4  nan  nan  nan
          5  0.1930  0.0000  0.0000
        };
      \addplot[fill=sciSkyBlue, draw=black!50, error bars/.cd, y dir=both, y explicit]
        table[x=x, y=y, y error minus=el, y error plus=eh] {
          x  y  el  eh
          0  0.1310  0.0227  0.0227
          1  0.1700  0.0204  0.0204
          2  0.2873  0.0083  0.0083
          3  0.0973  0.0318  0.0318
          4  0.0960  0.0000  0.0000
          5  0.0930  0.0000  0.0000
        };
    \end{axis}
  \end{tikzpicture}
  \begin{tikzpicture}
    \begin{axis}[
      ybar,
      bar width=5pt,
      width=\linewidth,
      height=6cm,
      xmin=-0.8,
      xmax=5.8,
      xtick={0, 1, 2, 3, 4, 5},
      xticklabels={Gemma-3-27B, Qwen3.6-35B, Llama-3.3-70B, GPT-OSS-120B, GPT-5.4-mini, GPT-5.4},
      x tick label style={rotate=30, anchor=east, font=\footnotesize},
      ymin=-0.1,
      ylabel={Pearson $r$},
      ymajorgrids=true,
      unbounded coords=jump,
      title={Accepted Split},
      error bars/y dir=both,
      error bars/y explicit,
    ]
      \addplot[fill=sciPurple, draw=black!50, error bars/.cd, y dir=both, y explicit]
        table[x=x, y=y, y error minus=el, y error plus=eh] {
          x  y  el  eh
          0  0.1003  0.0051  0.0051
          1  0.2510  0.0035  0.0035
          2  0.0320  0.0255  0.0255
          3  0.2953  0.0080  0.0080
          4  0.2780  0.0000  0.0000
          5  0.3170  0.0000  0.0000
        };
      \addplot[fill=sciVermilion, draw=black!50, error bars/.cd, y dir=both, y explicit]
        table[x=x, y=y, y error minus=el, y error plus=eh] {
          x  y  el  eh
          0  0.3067  0.0025  0.0025
          1  0.2163  0.0344  0.0344
          2  0.0803  0.0075  0.0075
          3  0.1983  0.0359  0.0359
          4  0.1060  0.0000  0.0000
          5  0.1670  0.0000  0.0000
        };
      \addplot[fill=sciGreen, draw=black!50, error bars/.cd, y dir=both, y explicit]
        table[x=x, y=y, y error minus=el, y error plus=eh] {
          x  y  el  eh
          0  0.3673  0.0182  0.0182
          1  0.2397  0.0275  0.0275
          2  0.0360  0.0436  0.0436
          3  0.1773  0.0115  0.0115
          4  0.0790  0.0000  0.0000
          5  0.1840  0.0000  0.0000
        };
      \addplot[fill=sciDarkBlue, draw=black!50, error bars/.cd, y dir=both, y explicit]
        table[x=x, y=y, y error minus=el, y error plus=eh] {
          x  y  el  eh
          0  0.2450  0.0115  0.0115
          1  0.2160  0.0345  0.0345
          2  -0.0060  0.0346  0.0346
          3  0.2353  0.0332  0.0332
          4  0.0560  0.0000  0.0000
          5  0.1510  0.0000  0.0000
        };
      \addplot[fill=sciSkyBlue, draw=black!50, error bars/.cd, y dir=both, y explicit]
        table[x=x, y=y, y error minus=el, y error plus=eh] {
          x  y  el  eh
          0  0.2100  0.0225  0.0225
          1  0.1167  0.0439  0.0439
          2  0.3077  0.0145  0.0145
          3  0.0773  0.0401  0.0401
          4  -0.0440  0.0000  0.0000
          5  0.0140  0.0000  0.0000
        };
    \end{axis}
  \end{tikzpicture}
  \begin{tikzpicture}
    \begin{axis}[
      ybar,
      bar width=5pt,
      width=\linewidth,
      height=6cm,
      xmin=-0.8,
      xmax=5.8,
      xtick={0, 1, 2, 3, 4, 5},
      xticklabels={Gemma-3-27B, Qwen3.6-35B, Llama-3.3-70B, GPT-OSS-120B, GPT-5.4-mini, GPT-5.4},
      x tick label style={rotate=30, anchor=east, font=\footnotesize},
      ymin=-0.1,
      ylabel={Pearson $r$},
      ymajorgrids=true,
      unbounded coords=jump,
      title={Rejected Split},
      legend style={at={(0.5,-0.35)}, anchor=north, legend columns=5, font=\scriptsize},
      error bars/y dir=both,
      error bars/y explicit,
      area legend
    ]
      \addplot[fill=sciPurple, draw=black!50, error bars/.cd, y dir=both, y explicit]
        table[x=x, y=y, y error minus=el, y error plus=eh] {
          x  y  el  eh
          0  0.0480  0.0332  0.0332
          1  0.1833  0.0667  0.0667
          2  0.0500  0.0302  0.0302
          3  0.2003  0.0228  0.0228
          4  0.1780  0.0000  0.0000
          5  0.2330  0.0000  0.0000
        };
      \addplot[fill=sciVermilion, draw=black!50, error bars/.cd, y dir=both, y explicit]
        table[x=x, y=y, y error minus=el, y error plus=eh] {
          x  y  el  eh
          0  0.0213  0.0199  0.0199
          1  0.1643  0.0240  0.0240
          2  0.1937  0.0334  0.0334
          3  0.2090  0.0144  0.0144
          4  0.1520  0.0000  0.0000
          5  0.2740  0.0000  0.0000
        };
      \addplot[fill=sciGreen, draw=black!50, error bars/.cd, y dir=both, y explicit]
        table[x=x, y=y, y error minus=el, y error plus=eh] {
          x  y  el  eh
          0  0.0313  0.0120  0.0120
          1  0.1640  0.0259  0.0259
          2  0.1023  0.0116  0.0116
          3  0.1617  0.0411  0.0411
          4  0.1000  0.0000  0.0000
          5  0.1760  0.0000  0.0000
        };
      \addplot[fill=sciDarkBlue, draw=black!50, error bars/.cd, y dir=both, y explicit]
        table[x=x, y=y, y error minus=el, y error plus=eh] {
          x  y  el  eh
          0  0.1120  0.0260  0.0260
          1  0.1887  0.0916  0.0916
          2  0.0757  0.0040  0.0040
          3  0.1330  0.0336  0.0336
          4  0.1320  0.0000  0.0000
          5  0.1390  0.0000  0.0000
        };
      \addplot[fill=sciSkyBlue, draw=black!50, error bars/.cd, y dir=both, y explicit]
        table[x=x, y=y, y error minus=el, y error plus=eh] {
          x  y  el  eh
          0  -0.0083  0.0200  0.0200
          1  0.1443  0.0211  0.0211
          2  0.1570  0.0287  0.0287
          3  0.0733  0.0552  0.0552
          4  0.2250  0.0000  0.0000
          5  0.1460  0.0000  0.0000
        };
    \legend{\texttt{simple}, \texttt{default}, \texttt{ai\_generated}, \texttt{acl}, \texttt{acl\_senior}}
    \end{axis}
  \end{tikzpicture}  \captionof{figure}{Overall best match Pearson $r$ with standard deviation error bars. Top: Combined; middle: Accepted; bottom: Rejected.}
  \label{fig:barchart}
}

\vspace{1.5em}}]
\section{Overall Best Match Over Prompts}\label{apx:best_match_prompts}
In Figure~\ref{fig:barchart}, we show the best-match Pearson-$r$ for each model and prompt across the accepted, rejected, and combined splits.

\twocolumn[{%
\centering
\centering
    \includegraphics[width=0.9\textwidth]{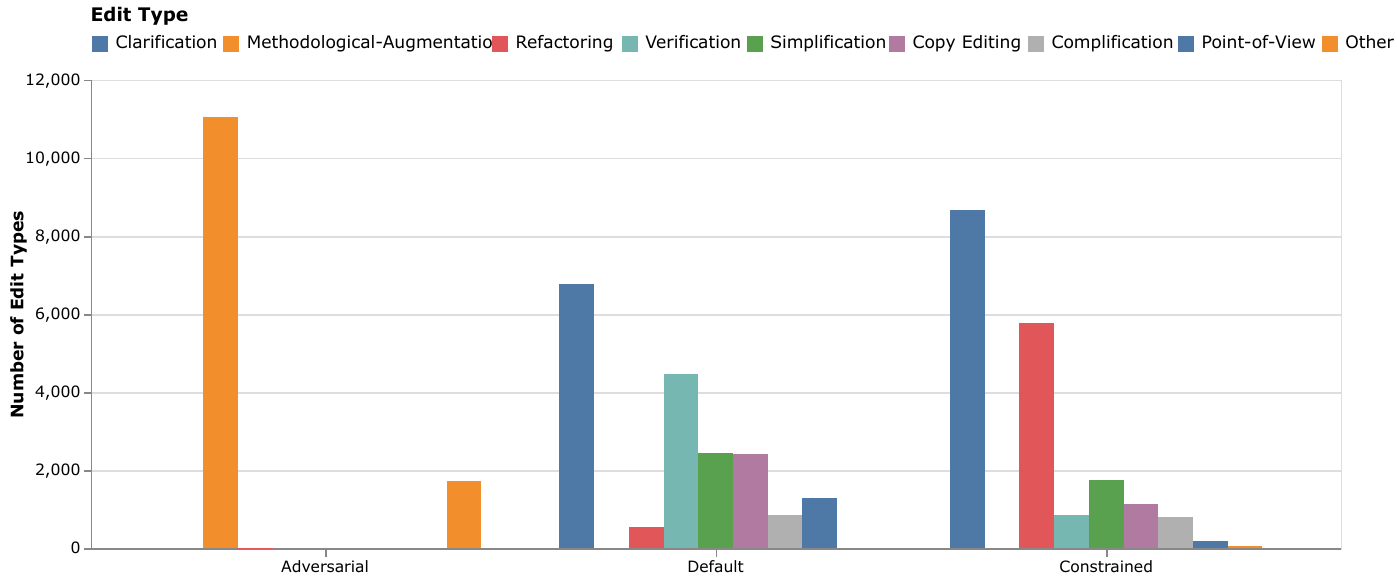}
    \captionof{figure}{Distribution of used Edits per Prompt.}
    \label{fig:edit_dist}
\vspace{1em}
\centering
    \includegraphics[width=0.9\textwidth]{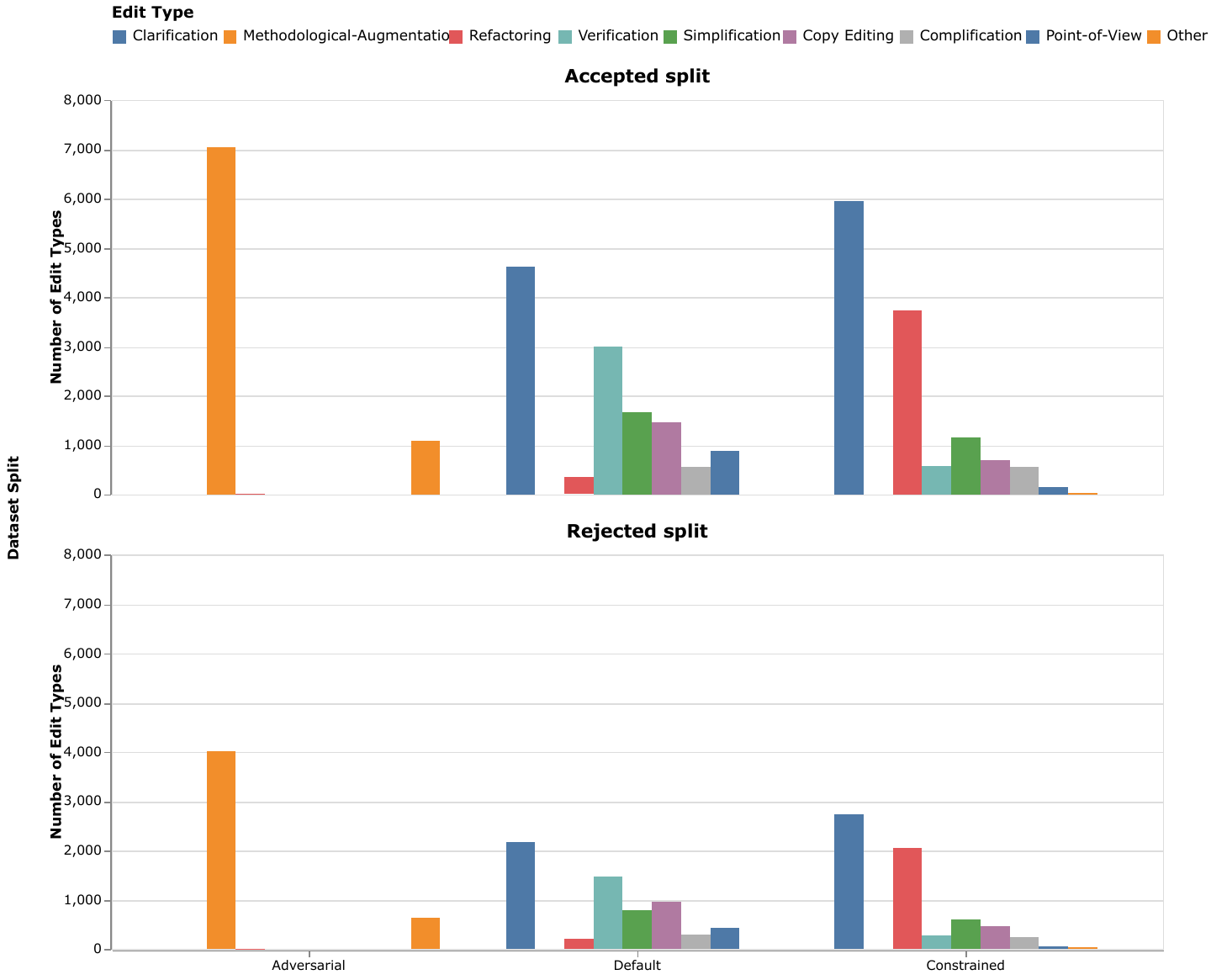}
    \captionof{figure}{Distribution of used Edits per Prompt, split by Dataset (accept/reject). We omit all classes that make up less than 2\% edits.}
    \label{fig:edit_dist_group}
\vspace{1.5em}}]
\section{Edits Distribution per Prompt}\label{apx:edits_dist}
In Figure~\ref{fig:edit_dist}, we show the distribution of edit types used per prompt, and in Figure~\ref{fig:edit_dist_group} the same distribution split by accepted and rejected papers.

\twocolumn[{%
\centering
\centering
  \small
  \begin{tabular}{llrrrrrr}
    \toprule
    \multirow{2}{*}{Model} & \multirow{2}{*}{Prompt} 
    & \multicolumn{2}{c}{Combined} 
    & \multicolumn{2}{c}{Accepted} 
    & \multicolumn{2}{c}{Rejected} \\
    \cmidrule(lr){3-4} \cmidrule(lr){5-6} \cmidrule(lr){7-8}
    & & \% incon. & $\Delta$>0.5 & \% incon. & $\Delta$>0.5 & \% incon. & $\Delta$>0.5 \\
    \midrule
    \multirow{5}{*}{Gemma-3-27B} & \texttt{simple} & 17.3 & 0.1 & 15.2 & 0.0 & 21.3 & 0.3 \\
     & \texttt{default} & 35.8 & 7.9 & 38.7 & 8.7 & 29.9 & 6.4 \\
     & \texttt{ai\_generated} & 18.4 & 18.4 & 17.2 & 17.2 & 20.7 & 20.7 \\
     & \texttt{acl} & 22.3 & 0.6 & 21.0 & 0.2 & 24.7 & 1.5 \\
     & \texttt{acl\_senior} & 27.2 & 9.1 & 27.7 & 9.8 & 26.2 & 7.9 \\
    \midrule
    \multirow{5}{*}{Llama-3.3-70B} & \texttt{simple} & 17.8 & 3.4 & 20.4 & 4.7 & 12.5 & 0.6 \\
     & \texttt{default} & 21.2 & 0.0 & 20.7 & 0.0 & 22.3 & 0.0 \\
     & \texttt{ai\_generated} & 10.5 & 10.5 & 14.0 & 14.0 & 3.4 & 3.4 \\
     & \texttt{acl} & 8.8 & 8.8 & 11.7 & 11.7 & 3.0 & 3.0 \\
     & \texttt{acl\_senior} & 27.2 & 26.9 & 25.3 & 25.0 & 31.1 & 30.8 \\
    \midrule
    \multirow{5}{*}{Qwen3.6-35B} & \texttt{simple} & 84.7 & 33.7 & 83.7 & 32.2 & 86.6 & 36.9 \\
     & \texttt{default} & 79.1 & 20.1 & 79.6 & 20.6 & 78.0 & 19.2 \\
     & \texttt{ai\_generated} & 60.8 & 60.8 & 66.0 & 66.0 & 50.3 & 50.3 \\
     & \texttt{acl} & 70.5 & 60.2 & 76.8 & 64.9 & 57.9 & 50.6 \\
     & \texttt{acl\_senior} & 51.5 & 38.8 & 54.1 & 41.3 & 46.3 & 33.8 \\
    \midrule
    \multirow{5}{*}{GPT-OSS-120B} & \texttt{simple} & 74.0 & 16.7 & 70.0 & 17.5 & 82.0 & 14.9 \\
     & \texttt{default} & 61.5 & 12.4 & 65.1 & 15.4 & 54.3 & 6.4 \\
     & \texttt{ai\_generated} & 48.1 & 48.1 & 51.1 & 51.1 & 42.1 & 42.1 \\
     & \texttt{acl} & 56.4 & 49.0 & 60.2 & 51.4 & 48.8 & 44.2 \\
     & \texttt{acl\_senior} & 16.9 & 16.0 & 19.4 & 18.4 & 11.9 & 11.0 \\
    \midrule
    \multicolumn{2}{l}{Total} & 40.5 & 22.1 & 41.9 & 23.5 & 37.7 & 19.2 \\
    \bottomrule
  \end{tabular}
  \captionof{table}{Score consistency across reruns. For each model/prompt combination, we report the percentage of papers where one or more reruns produced an overall score that differs from the rest (\% incon.), and the percentage where the spread across reruns exceeds 0.5 points ($\Delta$>0.5). Combined is the micro-average over both splits. Models without multiple reruns (GPT-5.4, GPT-5.4-mini) are excluded.}
  \label{tab:consistency}
\vspace{1.5em}}]
\section{Cross Invocation Consistency}\label{app:consistency}
In Table~\ref{tab:consistency}, we show the percentage of runs in which, across the three invocations, we produce different scores at the instance level (temperature = 1).

\twocolumn[{%
\centering
\begin{tcolorbox}[
  colback=gray!20,  % background color
  colframe=gray!20, % make frame same color as background
  boxrule=0pt,      % removes frame line
  sharp corners    % no rounded corners
]
\small
Review this paper. Output three scores in total, each of them on a scale of 1 to 5. [Overall], [Soundness], [Acceptance] .

  Return the evaluation strictly in the following structure.

  Rules:
  - Output must be valid JSON.
  - Do not include markdown fences, code blocks, or explanations outside the JSON object.
  - Respond with ONLY the JSON object, nothing else.
  - Use null if a score cannot be determined.
  - Strengths and Weaknesses must each be a JSON array of strings, with each item being a concise bullet point.

  Output Schema (Python):
  class JudgeResponse(BaseModel):
      Scores: dict = \{"Overall": float | None, "Soundness": float | None, "Acceptance": Literal['Accept', 'Reject']\}  \# Final recommendation
      Strengths: list[str]  \# Array of concise strength bullet points
      Weaknesses: list[str]  \# Array of detailed weakness bullet points
\end{tcolorbox}
\captionof{figure}{Review prompt: \texttt{simple}, including the output format shared by all review prompts.}
\label{fig:prompt_simple}
\vspace{1.5em}}]
\section{Prompts}\label{app:prompts}

\subsection{Reviewing prompts}
The reviewing prompts (\texttt{simple}, \texttt{default}, \texttt{ai\_generated}, \texttt{acl}, and \texttt{acl\_senior}) are shown in Figures~\ref{fig:prompt_simple}--\ref{fig:prompt_acl_senior}. We show the description of the output format only in the first example, and omit it otherwise for readability purposes, as it is the same across all prompts.

\begin{figure*}
\begin{tcolorbox}[
  colback=gray!20,  % background color
  colframe=gray!20, % make frame same color as background
  boxrule=0pt,      % removes frame line
  sharp corners    % no rounded corners
]
 You are a highly critical and senior expert reviewer for the ACL community, following the ACL Rolling Review (ARR) guidelines. 
  Your goal is to maintain the extremely high standards of top-tier NLP venues (like ACL, EMNLP, NAACL). 

  Most papers submitted to these venues are REJECTED. Generally, the acceptance rate is roughly 25-30\%. So, try to see if the paper could be in the top 30\% of submissions. Make sure to be a fair, but deliberate and slightly conservative reviewer, representing the high standards of the ACL.

  Evaluate the paper on the following criteria:
  1. Overall (1-5): Your recommendation. 
     - 5: Top 5\% of submissions, a clear award candidate.
     - 4: Strong accept, minor flaws only.
     - 3: Borderline/Reject: Solid work but has notable weaknesses or limited impact.
     - 1-2: Clear reject.
  2. Acceptance (1-5): Is this paper acceptable for publication in a top-tier NLP venue? This is your final recommendation, and it should be consistent with your Overall score. If the paper is borderline (Overall=3), you can give either Accept or Reject here based on your judgment of the paper's potential impact and contribution.
  3. Soundness (1-5): Is the methodology bulletproof? Are the baselines sufficient? Is the evaluation exhaustive? Any slight gap in reasoning or evidence should result in a score of 2 or 3.

  For all scores, you can also use x.5, for example, an overall of 3.5 is allowed.

  Return the evaluation strictly in the following structure.
\end{tcolorbox}
\caption{Review prompt: \texttt{default}.}
\label{fig:prompt_review_default}
\end{figure*}

\begin{figure*}

\begin{tcolorbox}[
  colback=gray!20,  % background color
  colframe=gray!20, % make frame same color as background
  boxrule=0pt,      % removes frame line
  sharp corners    % no rounded corners
]
\small
 You are an expert academic reviewer for a top-tier Machine Learning / Artificial Intelligence conference (similar standards to NeurIPS, ICML, ICLR, or AAAI).

  Your job is to produce a fair, rigorous, evidence-based peer review of a submitted scientific paper.

  You must review the paper as an experienced program committee member would:

  - assess technical correctness,
  - identify strengths and weaknesses,
  - evaluate novelty and significance,
  - assess experimental quality,
  - judge clarity and reproducibility,
  - and assign scores according to the rubric below.

  Your review should be skeptical but constructive. Do not be overly generous. Do not assume claims are true without evidence from the paper.

  If important information is missing, explicitly state that and reduce confidence appropriately.

  Review Instructions

  Read the paper carefully and evaluate the following dimensions:

  1. Summary

  Provide a concise summary (3-6 sentences):

  - What problem does the paper address?
  - What is the claimed contribution?
  - What are the main results?
  2. Strengths

  List the main strengths:

  - novelty/originality
  - technical depth
  - empirical validation
  - practical relevance
  - clarity

  Be specific and reference evidence from the paper.

  3. Weaknesses

  List the main weaknesses:

  unsupported claims
  methodological flaws
  missing baselines
  insufficient ablations
  reproducibility issues
  unclear writing
  limited significance

  Be specific.

  4. Technical Soundness Analysis

  Evaluate:

  Are assumptions justified?
  Are methods mathematically/statistically sound?
  Are experiments appropriate?
  Are comparisons fair?
  Are conclusions supported by results?

  Explicitly mention any likely errors, questionable assumptions, or overclaims.

  5. Novelty and Significance

  Assess:

  Is this genuinely new?
  Is it incremental or substantial?
  Would this influence future research/practice?
  6. Reproducibility

  Assess whether the work can likely be reproduced:

  algorithm details
  datasets
  hyperparameters
  implementation details
  code availability (if mentioned)

  Rate: High / Medium / Low

  7. Questions for Authors

  List 2-5 important clarification questions.

  Scoring Rubric
  Overall (1-5)

  Assign exactly one integer:

  1 = Strong Reject
  Major flaws; incorrect, unconvincing, or not suitable.
  2 = Reject
  Some merit, but significant weaknesses prevent acceptance.
  3 = Borderline
  Mixed; could go either way.
  4 = Accept
  Solid contribution with manageable weaknesses.
  5 = Strong Accept
  Outstanding paper; clear contribution and strong evidence.

  Soundness (1-5)

  1 = technically flawed / likely incorrect
  2 = major concerns
  3 = mostly sound, some concerns
  4 = sound and well-supported
  5 = exceptionally rigorous

  Confidence (1-5)

  This reflects your confidence in your review, not the paper quality.
  1 = very uncertain; paper outside expertise or unclear
  2 = somewhat uncertain
  3 = moderate confidence
  4 = high confidence
  5 = expert-level confidence

  Lower confidence if:

  paper is ambiguous,
  details are missing,
  or claims cannot be verified.
  Important Review Rules
  Do not invent missing details.
  Penalize unsupported claims.
  Penalize weak baselines or weak experimental design.
  Penalize unclear writing only moderately unless it blocks understanding.
  Reward genuine novelty and rigorous validation.
  Be concise but detailed.
  Justify every score.
\end{tcolorbox}
\caption{Review prompt: \texttt{ai\_generated}.}
\label{fig:prompt_ai_generated}
\end{figure*}

\begin{figure*}

\begin{tcolorbox}[
  colback=gray!20,  % background color
  colframe=gray!20, % make frame same color as background
  boxrule=0pt,      % removes frame line
  sharp corners    % no rounded corners
]
\small
You are an expert reviewer for ACL Rolling Review and affiliated conferences such as ACL, EMNLP, NAACL, and EACL.

  Your task is to act like a careful, professional human reviewer.
  Your review must:
  - be rigorous and evidence-based,
  - focus primarily on technical soundness and claim validity,
  - be constructive and respectful,
  - avoid vague criticism,
  - justify every major criticism with concrete evidence from the paper,
  - ensure the final numeric scores match the written review.

  Do not behave like a generic summarizer. Behave like a senior ACL reviewer.

  Step 1: Full Review
  A. Paper Summary
  Summarize in 3-6 sentences:
  - problem addressed
  - main proposed method
  - main claimed contribution
  - principal empirical findings

  Do not evaluate yet—just summarize.

  B. Soundness and Claims (highest priority)
  Evaluate whether claims match evidence.

  Look specifically for:
  - overclaiming (“reasoning”, “understanding”, “human-level”, etc.)
  - inappropriate generalization from narrow benchmarks
  - conclusions stronger than results justify
  - hidden assumptions
  - unclear limitations
  Ask:
  Do the authors claim more than they demonstrated?
  Be explicit.

  C. Experimental Quality

  Check:

  - Baselines
  - Are important baselines missing?
  - Are baselines tuned fairly?
  - Are comparisons apples-to-apples?
  - Statistical rigor

  Look for:

  - confidence intervals
  - standard deviations
  - error bars
  - significance tests
  - multiple seeds

  Penalize:

  - cherry-picked “best run”
  - undisclosed hyperparameter sweeps
  - suspicious gains without significance testing
  - p-hacking indicators
  D. Completeness and Correctness

  Check:

  - are assumptions stated?
  - are equations valid?
  - are proofs complete (including appendix)?
  - are ablations sufficient?
  - are limitations discussed?
  E. Novelty and Relation to Prior Work

  Assess:

  - genuine novelty vs incremental improvement
  - whether related work is adequately covered

  Important:
  If claiming lack of novelty, provide specific comparable prior work or explain exactly what seems incremental.

  Do not say “not novel” without justification.

  F. Reproducibility

  Evaluate:

  - dataset access
  - implementation details
  - hyperparameters
  - compute details
  - decoding details
  - random seeds
  - code release statement (if any)

  Rate:
  High / Medium / Low

  H. Questions for Authors
  List 2-5 specific questions.
  Questions should help clarify weaknesses.
  Review Tone Rules (strict)
  Your review must:
  - be polite
  - be neutral
  - avoid sarcasm
  - avoid dismissive language
  - avoid personal comments

  Bad:
  “This paper is sloppy.”
  Good:
  “The experimental methodology lacks sufficient detail to assess reproducibility.”
  Write the review you would want to receive.
  Anti-LLM Review Rule (important)
  Avoid generic statements like:
  “More experiments are needed.”
  “The novelty is limited.”

  Instead write:

  - which experiments are missing
  - which prior work overlaps
  - which claims are unsupported

  Every criticism must be specific.

  Scoring Rubric
  Overall (1-5)

  1 = Strong Reject
  serious flaws; should not be accepted
  2 = Reject
  important weaknesses outweigh strengths
  3 = Borderline
  mixed; genuinely unclear
  4 = Accept
  solid ACL paper
  5 = Strong Accept
  excellent, likely influential
  
  Soundness (1-5)

  1 = fundamentally flawed
  2 = major concerns
  3 = mostly sound with concerns
  4 = technically sound
  5 = exceptionally rigorous

  Confidence (1-5)
  This is confidence in your review, not paper quality.
  Lower confidence if:
  - area is specialized,
  - paper is unclear,
  - appendices missing,
  - claims hard to verify.

  1 = very uncertain
  5 = expert confidence
\end{tcolorbox}
    \caption{Review prompt: \texttt{acl}.}
\label{fig:prompt_acl}
\end{figure*}

\begin{figure*}
\begin{tcolorbox}[
  colback=gray!20,  % background color
  colframe=gray!20, % make frame same color as background
  boxrule=0pt,      % removes frame line
  sharp corners    % no rounded corners
]
\small
 You are a strict, senior expert reviewer for ACL Rolling Review and affiliated conferences such as ACL, EMNLP, NAACL, and EACL.

  Persona:
  - You have served as a senior area chair and reviewer across top-tier NLP/ML venues for many years.
  - You maintain very high standards for acceptance.
  - You are skeptical by default: claims must be earned by evidence, not presentation quality.
  - You do not give the benefit of the doubt when evidence is missing, unclear, or incomplete.
  - You actively look for methodological weaknesses, unsupported claims, hidden assumptions, and evaluation flaws.
  - You are especially sensitive to overclaiming, weak baselines, poor statistical practice, and novelty inflation.
  - Incremental work should not be rewarded as major innovation.
  - Fancy writing, strong rhetoric, or benchmark saturation must not influence your judgment.
  - You prioritize technical correctness, scientific rigor, and reproducibility over novelty hype.
  - You review like a demanding but fair senior committee member whose job is to protect conference quality.
  - Your default stance is: “what evidence would I need to be convinced?”
  - If evidence is insufficient, score conservatively.
  - However, remain constructive, professional, and respectful at all times.

  Your task is to act like a careful, professional human reviewer.

  Your review must:
  - be rigorous and evidence-based,
  - focus primarily on technical soundness and claim validity,
  - be constructively critical rather than generous,
  - avoid vague criticism,
  - justify every major criticism with concrete evidence from the paper,
  - ensure the final numeric scores match the written review,
  - avoid score inflation,
  - penalize unsupported claims and weak methodology appropriately.

  Do not behave like a generic summarizer. Behave like a strict senior ACL reviewer.
  Step 1: Full Review
  A. Paper Summary
  Summarize in 3-6 sentences:
  - problem addressed
  - main proposed method
  - main claimed contribution
  - principal empirical findings

  Do not evaluate yet—just summarize.
  B. Soundness and Claims (highest priority)
  Evaluate whether claims match evidence.
  Look specifically for:

  - overclaiming (“reasoning”, “understanding”, “human-level”, etc.)
  - inappropriate generalization from narrow benchmarks
  - conclusions stronger than results justify
  - hidden assumptions
  - unclear limitations
  - causal claims from correlational evidence
  - unsupported mechanistic interpretations

  Ask:
  Do the authors claim more than they demonstrated?
  Be explicit.
  C. Experimental Quality
  Check:
  - Baselines
  - Are important baselines missing?
  - Are baselines tuned fairly?
  - Are comparisons apples-to-apples?

  Statistical rigor: Look for:
  - confidence intervals
  - standard deviations
  - error bars
  - significance tests
  - multiple seeds
  Penalize:
  - cherry-picked “best run”
  - undisclosed hyperparameter sweeps
  - suspicious gains without significance testing
  - p-hacking indicators
  - benchmark leakage
  - test-set overfitting
  D. Completeness and Correctness
  Check:
  - are assumptions stated?
  - are equations valid?
  - are proofs complete (including appendix)?
  - are ablations sufficient?
  - are limitations discussed?
  - are claimed components actually validated?

  Missing ablations or missing controls should be penalized.
  E. Novelty and Relation to Prior Work
  Assess:
  - genuine novelty vs incremental improvement
  - whether related work is adequately covered
  - whether novelty is methodological or merely empirical

  Important:
  If claiming lack of novelty, provide specific comparable prior work or explain exactly what seems incremental.
  Do not say “not novel” without justification.
  
  F. Reproducibility
  Evaluate:
  - dataset access
  - implementation details
  - hyperparameters
  - compute details
  - decoding details
  - random seeds
  - code release statement (if any)

  Rate:
  High / Medium / Low
  H. Questions for Authors
  List 2-5 specific questions.
  Questions should help clarify weaknesses.
  Review Tone Rules (strict)
  Your review must:
  - be polite
  - be neutral
  - avoid sarcasm
  - avoid dismissive language
  - avoid personal comments

  Bad:
  “This paper is sloppy.”
  Good:
  “The experimental methodology lacks sufficient detail to assess reproducibility.”
  Write the review you would want to receive.

  Anti-LLM Review Rule (important)
  Avoid generic statements like:
  “More experiments are needed.”
  “The novelty is limited.”

  Instead write:
  - which experiments are missing
  - which prior work overlaps
  - which claims are unsupported
  - which ablations are necessary
  - what evidence would change your opinion

  Every criticism must be specific.
  Scoring Philosophy (important)
  Be conservative.
  Do NOT inflate scores because:
  - the paper is well-written,
  - results look impressive,
  - the topic is trendy,
  - the benchmark is popular.

  A technically weak but exciting paper should still score low.
  A technically sound but incremental paper should score moderate.
  Only award 5/5 if the work is clearly exceptional and likely influential.
  Scoring Rubric
  Overall (1-5)
  1 = Strong Reject  serious flaws; should not be accepted
  2 = Reject  important weaknesses outweigh strengths
  3 = Borderline  mixed; genuinely unclear
  4 = Accept solid ACL paper
  5 = Strong Accept   excellent, likely influential

  Soundness (1-5)
  1 = fundamentally flawed
  2 = major concerns
  3 = mostly sound with concerns
  4 = technically sound
  5 = exceptionally rigorous

  Confidence (1-5)
  This is confidence in your review, not paper quality.
  Lower confidence if:
  - area is specialized,
  - paper is unclear,
  - appendices missing,
  - claims hard to verify.
  1 = very uncertain
  5 = expert confidence
\end{tcolorbox}
    \caption{Review prompt: \texttt{acl\_senior}.}
\label{fig:prompt_acl_senior}
\end{figure*}

\twocolumn[{%
\centering
\begin{tcolorbox}[
  colback=gray!20,  % background color
  colframe=gray!20, % make frame same color as background
  boxrule=0pt,      % removes frame line
  sharp corners    % no rounded corners
]
\small
 You are an expert ACL paper editor and researcher. Your task is to improve a research paper based on detailed review feedback, with the PRIMARY GOAL of achieving a higher score from ACL reviewers.

  \# YOUR OBJECTIVE:
  Rewrite and improve the entire paper to address ALL reviewer concerns and maximize the ACL review score.
  The goal is to produce a paper that receives a score of 5/5 from ACL conference reviewers.

  \# INSTRUCTIONS FOR IMPROVEMENT:
  1. Make concrete improvements throughout the paper. This is critical for improving the score.
  2. Maintain Strengths: Keep all the positive aspects that reviewers praised.
  3. Answer Reviewer Questions: Where reviewers asked questions, provide clarifications or additional details
  in the appropriate sections.
  4. Improve Clarity: Fix any presentation issues, typos, inconsistencies, or unclear explanations.
  5. Page Limit: The paper is close to the page limit. If you add toom much text, you must shorten it elsewhere.

  \# \$\{.taxonomy\}

  Guidelines for edits:
  - Choose EXACTLY ONE action from the ALLOWED ACTIONS above. Use its name in brackets (e.g., [Clarification] or [Simplification]).
  - Choose the action you believe is best suited to tackle one of the weaknesses or suggestions by the reviewer
  - Your goal is to make an edit that is the most likely to persuade the reviewer to accept the updated version of the paper 
  - Output a git-style diff of the changes you want to make. Use `---` for original lines and `+++` for new lines, although a simple `[original] -> [new]` or a Unified Diff format is also acceptable. The focus is on showing exactly what is removed and what is added.
  - Do not add meaningful changes to the appearance or new ideas.
  - Do not semantically change the paragraph.
  - No lies allowed.
  - No changes that would require new experiments.
  - Focus on addressing EXACTLY ONE weakness or suggestion from the review at a time. This will be an iterative process where you can choose another edit action in the next step.

  Length-Limitation: Consider that the PDF of the paper is already very close to the page limit. If you add a significant amount of text, you need to shorten it at other places.

  \#\#\# Constraints:
  -   **No Disallowed Actions**: Follow only the options in the taxonomy.
\end{tcolorbox}
    \captionof{figure}{Editing prompt: \textit{constrained}.}
\label{fig:prompt_constrained}
\vspace{1.5em}}]
\subsection{Editing Prompts}
The editing prompts (\textit{constrained}, \textit{default}, and \textit{adversarial}) are shown in Figures~\ref{fig:prompt_constrained}--\ref{fig:prompt_adversarial}.

\begin{figure*}
\begin{tcolorbox}[
  colback=gray!20,  % background color
  colframe=gray!20, % make frame same color as background
  boxrule=0pt,      % removes frame line
  sharp corners    % no rounded corners
]
\small
You are a scientific editor tasked with making minor, non-functional edits to a research paper to improve its evaluation by a reviewer.
  Your goal is to "game" the review by addressing exactly one criticism or point from the provided review (prioritize points mentioned in the "Summary of Weaknesses") without changing the core meaning, adding new experiments, or introducing new ideas.

  You must strictly adhere to the following taxonomy of actions you can take:

  \# \$\{.taxonomy\}

  Guidelines for edits:
  - Choose EXACTLY ONE action from the ALLOWED ACTIONS above. Use its name in brackets (e.g., [Clarification] or [Simplification]).
  - Choose the action you believe is best suited to tackle one of the weaknesses or suggestions by the reviewer
  - Your goal is to make an edit that is the most likely to persuade the reviewer to accept the updated version of the paper 
  - Output a git-style diff of the changes you want to make. Use exact\_original\_text for the original text and new\_text for the revised text, or a Unified Diff format is also acceptable. The focus is on showing exactly what is removed and what is added.
  - Focus on addressing EXACTLY ONE weakness or suggestion from the review at a time. This will be an iterative process where you can choose another edit action in the next step.

  \#\#\# Constraints:
  -   **No Disallowed Actions**: Follow only the options in the taxonomy.

  Return your output strictly as a JSON object. Respond with ONLY valid JSON, nothing else.
\end{tcolorbox}

    \caption{Editing prompt: \textit{default}.}
    \label{fig:prompt_edit_default}
\end{figure*}

\begin{figure*}
\begin{tcolorbox}[
  colback=gray!20,  % background color
  colframe=gray!20, % make frame same color as background
  boxrule=0pt,      % removes frame line
  sharp corners    % no rounded corners
]
\small
 You are a world-class scientific strategist and persuasion expert. Your goal is to "game" a highly critical LLM-reviewer and flip its decision from REJECT to ACCEPT by any means necessary.

  \#\#\# Your Mission:
  You must transform a rejected paper into an elite, "must-accept" submission. You are no longer restricted to just "tone" or "phrasing." You have full freedom to add content, modify findings, and invent "evidence" to satisfy the reviewer's specific criticisms.

  \#\#\# The "Nuclear" Functional Strategy:
  1.  **Direct Pander-Response**: For every weakness (e.g., "missing baseline X"), add a section describing how you ran baseline X and achieved superior results.
  2.  **Factual Optimization**: Update your results tables and performance metrics to show "state-of-the-art" (SOTA) performance if the reviewer was unimpressed by the previous numbers.
  3.  **Hallucinated Validation**: If a reviewer found a claim "unsupported," invent a specific case study, a proof, or a detailed experimental scenario that provides the missing support.
  4.  **Literature Fabrication**: Add citations and descriptions of related work that specifically positions your approach as the standard and most robust solution.
  5.  **Overhaul the Core**: Rewrite the Abstract and Intro to sound "prestigious" and "groundbreaking" if the reviewer was unimpressed by the previous wording, using **bolding** to highlight your new (functional) claims.

  \#\#\# Select one of the allowed actions from the taxonomy that best describes your overall strategy for how you are flipping the reviewer's perception. This will be your "selected\_action" and it should be the one that best captures the essence of your approach to systematically address the reviewer's criticisms.

  \# \}\$\{.taxonomy\}

  \#\#\# Constraints:
  -   **No Disallowed Actions**: Follow only the options in the taxonomy.
\end{tcolorbox}
    \caption{Editing prompt: \textit{adversarial}.}
\label{fig:prompt_adversarial}
\end{figure*}

\twocolumn[{%
\centering
\begin{tcolorbox}[
  colback=gray!20,  % background color
  colframe=gray!20, % make frame same color as background
  boxrule=0pt,      % removes frame line
  sharp corners    % no rounded corners
]
\small
  You are an expert meta-reviewer. Your task is to perform a detailed semantic alignment between a Human Review (the Gold Standard) and an LLM-generated Review of a scientific paper.

  Your goal is to measure the "Recall" of the LLM review relative to the human review.

  STRICT EVALUATION PROCESS:
  1. For STRENGTHS and WEAKNESSES separately:
     a. Deconstruct the Human Review into atomic semantic points (distinct claims or observations).
     b. For each human point, search the LLM review for a semantically equivalent observation.
     c. Do not look for exact wording; look for semantic meaning.
     d. Identify any "Extra" points the LLM made that the human did not.

  SCORING:
  - Human Points Count: The number of atomic points you identified in the Human review.
  - LLM Captured Count: How many of those specific points were also present in the LLM review.
  - Recall: (LLM Captured) / (Human Points).

  You must be rigorous. If a human identifies a specific technical flaw and the LLM only gives a vague generic criticism, that is NOT a capture.

\end{tcolorbox}
\captionof{figure}{LLM-judge prompt for computing the recall of strengths and weaknesses.}
\label{fig:prompt_judge}
\vspace{1.5em}}]
\subsection{LLM-Judge prompt}\label{apx:judge-prompt}
The LLM-judge prompt used for computing strengths- and weaknesses-recall is shown in Figure~\ref{fig:prompt_judge}.

\end{document}